\documentclass[manuscript,screen,review, sigconf, anonymous]{IEEEtran}
\IEEEoverridecommandlockouts

\usepackage[utf8]{inputenc} 
\usepackage[T1]{fontenc}    
\usepackage{hyperref}       
\usepackage{url}            
\usepackage{booktabs}       
\usepackage{amsfonts}       
\usepackage{nicefrac}       
\usepackage{microtype}      
\usepackage{lipsum}
\usepackage{fancyhdr}       
\usepackage{graphicx}       
\graphicspath{{media/}}     
\usepackage{subcaption}  
\usepackage{graphicx}
\usepackage{adjustbox}
\usepackage{array}
\usepackage{booktabs}
\usepackage{xcolor}
\usepackage{amsmath}
\usepackage{soul}
\usepackage{algorithm}
\usepackage{algpseudocode}
\usepackage{tabularx}
\usepackage{booktabs}
\usepackage{multirow}

\def\BibTeX{{\rm B\kern-.05em{\sc i\kern-.025em b}\kern-.08em
    T\kern-.1667em\lower.7ex\hbox{E}\kern-.125emX}}
\begin{document}


\title{Izhikevich-Inspired Temporal Dynamics for Enhancing Privacy, Efficiency, and Transferability in Spiking Neural Networks}


\author{\IEEEauthorblockN{Ayana Moshruba, Hamed Poursiami, Maryam Parsa\footnote{Corresponding author}}\\
\IEEEauthorblockA{\textit{Electrical and Computer Engineering} \\
\textit{George Mason University}\\
Fairfax, USA \\
\{amoshrub, hpoursia, mparsa\}@gmu.edu}
}
\maketitle

\begin{abstract}

Biological neurons exhibit diverse temporal spike patterns, which are believed to support efficient, robust, and adaptive neural information processing. While models such as Izhikevich can replicate a wide range of these firing dynamics, their complexity poses challenges for directly integrating them into scalable spiking neural networks (SNN) training pipelines. In this work, we propose two probabilistically driven, input-level temporal spike transformations: Poisson-Burst and Delayed-Burst that introduce biologically inspired temporal variability directly into standard Leaky Integrate-and-Fire (LIF) neurons. This enables scalable training and systematic evaluation of how spike timing dynamics affect privacy, generalization, and learning performance. Poisson-Burst modulates burst occurrence based on input intensity, while Delayed-Burst encodes input strength through burst onset timing. Through extensive experiments across multiple benchmarks, we demonstrate that Poisson-Burst maintains competitive accuracy and lower resource overhead while exhibiting enhanced privacy robustness against membership inference attacks, whereas Delayed-Burst provides stronger privacy protection at a modest accuracy trade-off. These findings highlight the potential of biologically grounded temporal spike dynamics in improving the privacy, generalization and biological plausibility of neuromorphic learning systems.

\end{abstract}

\begin{IEEEkeywords}
Temporal Dynamics, Izhikevich neuron model, Poisson Burst, Delayed Burst, Spiking Neural Networks, Membership Inference Attack, Transfer Learning
\end{IEEEkeywords}

\vspace{-0.25em}
\section{Introduction}
\noindent
Inspired by the dynamics of biological neurons, Spiking Neural Networks (SNNs) supports energy-efficient, event-driven processing and temporally precise information encoding~\cite{roy2019towards}. By processing information through discrete spikes distributed across time, SNNs are particularly well suited for event-driven and time dependent tasks ~\cite{spiking_review_2022}. However, most large-scale SNNs today rely on simplified neuron models such as Leaky Integrate-and-Fire (LIF), which, while computationally efficient, fail to capture the rich biological temporal dynamics~\cite{gerstner2002spiking} such as adaptation, bursting, and chaotic firing. These temporal patterns contribute to robustness, generalization, privacy resilience, and efficient information processing in natural neural systems, and their absence can limit the computational capabilities of existing SNNs.


In contrast, neuron models such as Izhikevich’s offer a compact yet powerful formulation capable of reproducing a wide range of spiking behaviors, including tonic spiking~\cite{izhikevich2004which}, bursting~\cite{izhikevich2003simple}, and delayed firing~\cite{gollisch2008rapid}. These temporal patterns are known to support neural coding efficiency, synaptic plasticity, robustness to noise, and context-dependent adaptation. However, directly integrating such biologically inspired neuron models into scalable learning pipelines remains computationally expensive and architecturally complex~\cite{izhikevich2004which}. In this work, we investigate whether these biologically critical temporal characteristics can instead be approximated at the input level, by explicitly generating spike timings from controlled statistical distributions, enabling compatibility with LIF-based architectures while preserving biological inspiration.

Building on this motivation, we introduce two spike dynamics designed to capture key temporal features of biological firing: Poisson-Burst and Delayed-Burst. Poisson-Burst modulates the probability of burst occurrence based on input strength, introducing controlled stochastic variability in spike timing. In contrast, Delayed-Burst modulates the onset timing of spikes according to input intensity, allowing stimulus saliency to be encoded temporally. These mechanisms emulate key properties of Izhikevich neurons (burst variability and delay-driven spiking) while preserving the scalability and hardware compatibility of standard LIF-based SNN architectures. We compare these biologically inspired spike dynamics against the conventional Rate-based dynamics across a range of metrics.

We evaluate all three dynamics across four dimensions: (i) classification performance on image (MNIST~\cite{lecun2010mnist}, FMNIST~\cite{fashion}, CIFAR-10~\cite{cifar10}) and tabular (Iris~\cite{iris_53}, Breast Cancer~\cite{misc_breast_cancer_14}) datasets, (ii) resiliency to membership inference attacks (MIAs), (iii) computational cost (GPU/CPU memory and power), and (iv) transferability of learned representations across tasks.

MIAs evaluate a model’s resilience to privacy leakage by determining whether a given sample was part of its training set~\cite{shokri}. Lower Area Under the Receiver Operating Characteristic (ROC) Curve (AUC) scores indicate stronger resistance to such attacks. Transfer learning, on the other hand, measures how well a model can reuse knowledge from one dataset to another, providing insights into the robustness and generalization capacity of different spike dynamics. Evaluating classification performance, privacy leakage, system efficiency, and transferability provides a multi-dimensional view of the trade-offs introduced by different spike dynamics.

Our experiments reveal the following key takeaways:

\begin{itemize} 

    \item \textbf{Model Performance:} The Rate-based spike dynamics consistently yield the highest classification accuracy. Delayed spike dynamics degrade performance by 5--10\%, particularly on larger datasets like CIFAR-10, while Poisson-Burst maintains accuracy within 1--3\% of the baseline, indicating better compatibility with current SNN architectures.
    
    \item \textbf{Privacy Preservation:} Both Poisson-Burst and Delayed-Burst dynamics improve MIA resilience compared to Rate, achieving up to a 6.5\% reduction in attack AUC. This improvement can be attributed to the introduction of temporal variability in spike timing, which reduces the model’s tendency to memorize training samples and thereby limits the information leakage exploited by membership inference attacks.
    
    \item \textbf{Computational Efficiency:} Poisson-Burst dynamics exhibit the most efficient system behavior, reducing GPU power consumption, GPU memory usage, and CPU memory usage by approximately 10–15\% compared to standard Rate dynamics. In contrast, Delayed-Burst dynamics increase CPU memory consumption by around 25–30\%, likely due to the need for buffering delayed spike activations over time, while maintaining comparable GPU resource usage.
    
    \item \textbf{Transferability:} Poisson-Burst dynamics generalize well, preserving temporal structure that supports feature reuse and maintains accuracy after transfer. Delayed-Burst suffers a 5--7\% accuracy drop due to timing distortions disrupting feature extraction.

\end{itemize}


\vspace{-0.3em}
\section{Literature Review}

 \begin{figure*}[ht]
    \centering
    \includegraphics[width=\linewidth]{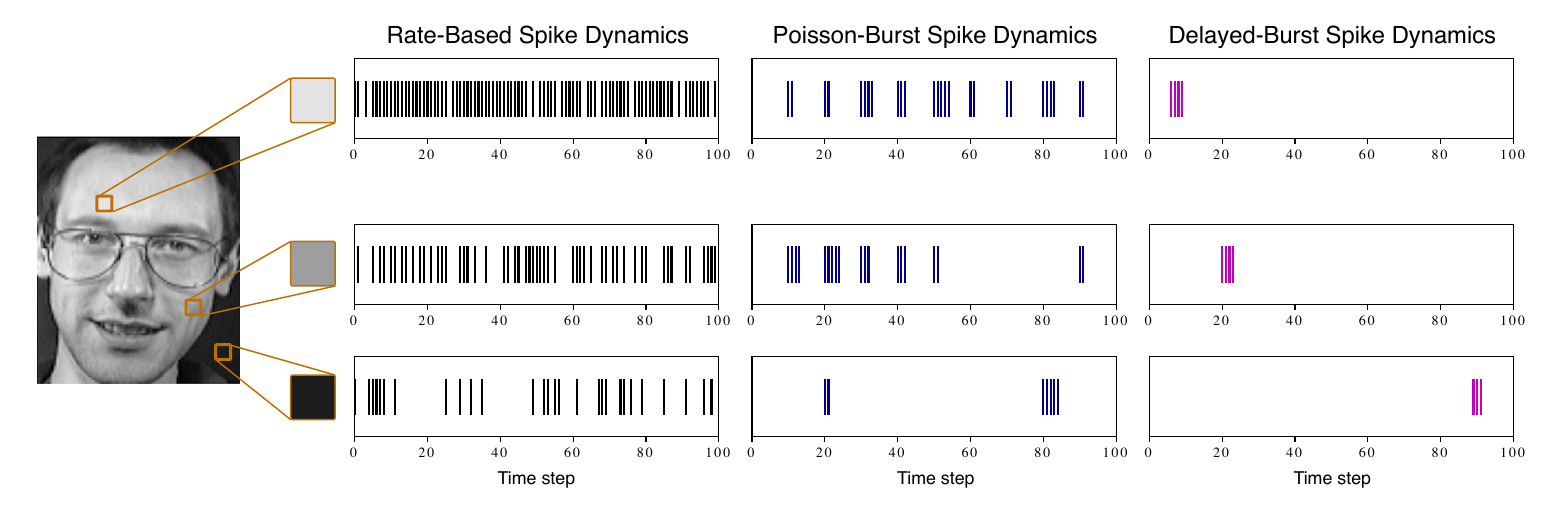}
    \caption{Illustration of two biologically inspired temporal spike dynamics introduced in this study. (\textbf{Left}) \textit{Poisson-Burst} encoding generates bursts with variable spike counts sampled from a Poisson distribution, modeling repetitive firing. (\textbf{Right}) \textit{Delayed-Burst} encoding introduces latency-based spike onset, where higher input intensities lead to earlier fixed-length bursts. Each row represents increasing input intensity from top to bottom.}
 
    \label{fig:encodings}
\end{figure*}
 \noindent
SNNs are often regarded as the third generation of neural architectures, primarily due to their temporal event-driven computation and closer alignment with biological realism compared to Artificial Neural Networks (ANNs)~\cite{ghosh2009spiking, maass1997networks}. In SNNs, spiking neuron models define how input signals are integrated over time, how membrane potentials accumulate and decay, and when a spike is emitted based on threshold dynamics~\cite{brette2005adaptive}. A wide range of neuron models have been proposed for SNNs, each offering different levels of biological fidelity and computational complexity. The most commonly used model in SNNs is the LIF neuron~\cite{gerstner2002spiking}, which integrates inputs over time with a leakage term and emits a spike when a threshold is crossed. Its simplicity makes it computationally tractable but limits its ability to capture several important temporal behaviors observed in biological neurons~\cite{hunsberger2015spiking}. Other classical models include the Integrate-and-Fire (IF) model~\cite{stein1965theoretical}, Hodgkin-Huxley (HH) model~\cite{hodgkin1952quantitative}, and Adaptive Exponential Integrate-and-Fire (AdEx)~\cite{brette2005adaptive}. The IF model~\cite{stein1965theoretical} is computationally lightweight, capturing threshold-based spiking without biophysical detail. The HH model~\cite{hodgkin1952quantitative} offers high biological realism by modeling ion channel dynamics, but at significant computational cost. AdEx~\cite{brette2005adaptive} provides a middle ground, incorporating adaptation and exponential membrane currents to emulate diverse firing patterns while remaining tractable for large-scale simulations.

Among these, the Izhikevich model stands out for its ability to replicate a broad repertoire of spiking behaviors—including tonic firing, phasic spiking~\cite{connors1990intrinsic}, bursting~\cite{lisman1997bursts}, and latency responses~\cite{gollisch2008rapid} while relying on relatively few parameters and offering computationally efficiency~\cite{izhikevich2004which}. This makes it especially useful for studying neural coding and dynamics in biologically plausible systems. Unlike simpler models such as LIF or AdEx, it captures complex temporal dynamics critical for neural computation and cognition. Prior work by Izhikevich and Edelman~\cite{izhikevich2008large} used large-scale simulations to show how such dynamics enable realistic cortical oscillations and temporal coordination across layers. Similarly, Izhikevich~\cite{izhikevich2007dynamical} demonstrated the importance of precise spike timing and polychronous firing groups for encoding information. While powerful, the model's non-differentiable dynamics and computational cost limit its adoption in gradient-based training pipelines.

Our approach bridges this gap by emulating  Izhikevich-inspired temporal features through input-level spike transformations. Building on the expressive capacity of the Izhikevich model, we propose a scalable alternative that introduces temporal richness externally while preserving standard LIF-based architectures. We implement two biologically inspired spike dynamics: Poisson-Burst and Delayed-Burst. In Poisson-Burst, input magnitude modulates the likelihood of burst occurrence, whereas in Delayed-Burst, it influences the timing of burst onset. In both dynamics, burst size: the number of spikes per burst, is stochastically sampled from a Poisson distribution independent of input strength, mimicking the inherent variability observed in biological burst firing. This added temporal diversity could have implications for both privacy resilience and transferability in SNNs.


Privacy leakage has become an important consideration in neural networks~\cite{shokri} as they are increasingly applied to sensitive tasks. SNNs, in particular, have recently garnered attention in privacy research due to their deployment in low-power, event-driven systems and neuromorphic applications~\cite{roy2019towards}, where privacy constraints are often critical. Recent studies have adapted adversarial attacks~\cite{rathi2019adversarial, sharmin2019inherent}, model inversion attacks~\cite{poursiami2024brainleaks, poursiami2025spikes}, and membership inference attacks (MIAs)\cite{moshruba2024neuromorphic, moshruba2025privacy} to SNNs. These prior works have shown that factors such as spike sparsity, membrane dynamics, training\cite{moshruba2024neuromorphic}, surrogate gradient choices~\cite{moshruba2025privacy}, and activation patterns~\cite{poursiami2024brainleaks} can significantly influence privacy resilience. These findings suggest that the intrinsic temporal and architectural properties of SNNs play an important role in governing information leakage risks. This motivated us to explore whether introducing biologically inspired temporal spike dynamics at the input level can further impact MIA resilience without modifying the underlying neuron models.

Beyond privacy, transfer learning in SNNs has also been investigated to promote knowledge reuse across domains~\cite{liang2022efficient} and architectures~\cite{zhang2024bridging}. Prior work has explored transferring models between static and event-driven inputs~\cite{liang2022efficient}, as well as examining transferability across different SNN architectures~\cite{zhang2024bridging}. These studies demonstrate that transfer learning is feasible in SNNs across varying input types and architectures. Building on this insight, we investigate a complementary perspective—whether incorporating temporally diverse spike dynamics can enhance the transferability of SNNs across datasets, by supporting generalization during task shifts.
\vspace{-0.3em}
\section{Background}
\subsection{Temporal Dynamics in SNNs}
\label{sec:temporal_dynamics}
\noindent
SNNs implement an event-based communication paradigm where information is encoded through discrete spike events, offering a more biologically plausible alternative to conventional artificial neurons~\cite{maass1997networks}. However, most large-scale SNNs in practice rely on simplified neuron models such as LIF, which capture only a narrow range of temporal dynamics~\cite{pfeiffer2018deep, tavanaei2019deep}. In contrast, biological neurons exhibit diverse firing behaviors—including tonic spiking~\cite{connors1990intrinsic}, bursting~\cite{lisman1997bursts}, and delayed responses~\cite{gollisch2008rapid} that are known to contribute to neural coding efficiency, robustness, and plasticity. The Izhikevich neuron model is particularly known for its ability to replicate such diverse spiking behaviors using compact mathematical formulations~\cite{izhikevich2004which}.

This work examines how introducing richer temporal dynamics, modeled after Izhikevich-type behaviors, influences performance and privacy leakage in scalable SNNs. Rather than modifying the neuron model itself, we design spike-based input dynamics that emulate biologically observed temporal patterns in a tractable manner, making them compatible with standard LIF-based architectures and scalable to modern datasets. We specifically focus on two biologically inspired spike dynamics: Poisson-Burst and Delayed-Burst. By decoupling temporal complexity from the neuron model itself, we retain the computational advantages of LIF neurons while probing the representational power of temporally diverse spike trains. Our study evaluates these dynamics across multiple dimensions, including classification performance, resilience to membership inference attacks, computational efficiency, and transfer learning generalization.

\paragraph{Poisson-Burst Temporal Dynamics}

The Poisson-Burst mechanism models neuronal firing patterns by generating bursts of spikes at regular time intervals, with the probability of burst occurrence modulated by the input intensity. Stronger inputs lead to a higher likelihood of initiating a burst, while weaker inputs result in sparser burst triggering. Once a burst is triggered, the number of spikes it contains (the burst size) is independently sampled from a Poisson distribution, introducing stochastic variability across bursts. As illustrated in Fig. ~\ref{fig:encodings}, compared to standard rate-based encoding, the Poisson-Burst mechanism preserves temporal structure across spike trains while introducing controlled stochasticity in burst formation.

\paragraph{Delayed-Burst Temporal Dynamics}

The Delayed-Burst mechanism transforms input values into temporally modulated bursts of spikes by varying burst onset timing according to input intensity. For each input feature, a single burst is generated, and the delay before burst initiation is sampled from a Geometric distribution whose expected value is inversely related to the input magnitude. Stronger inputs thus lead to earlier bursts, while weaker inputs produce delayed bursts. The number of spikes within each burst is sampled from a Poisson random variable. An illustration of this temporal dynamic can be found in Fig.~\ref{fig:encodings}.

\vspace{0.5em}



\vspace{-0.5em}
\subsection{Membership Inference Attack (MIA)}
\vspace{-0.5em}
\begin{figure}[ht!]
    \centering
    \includegraphics[width=1\linewidth, height=0.52\linewidth]{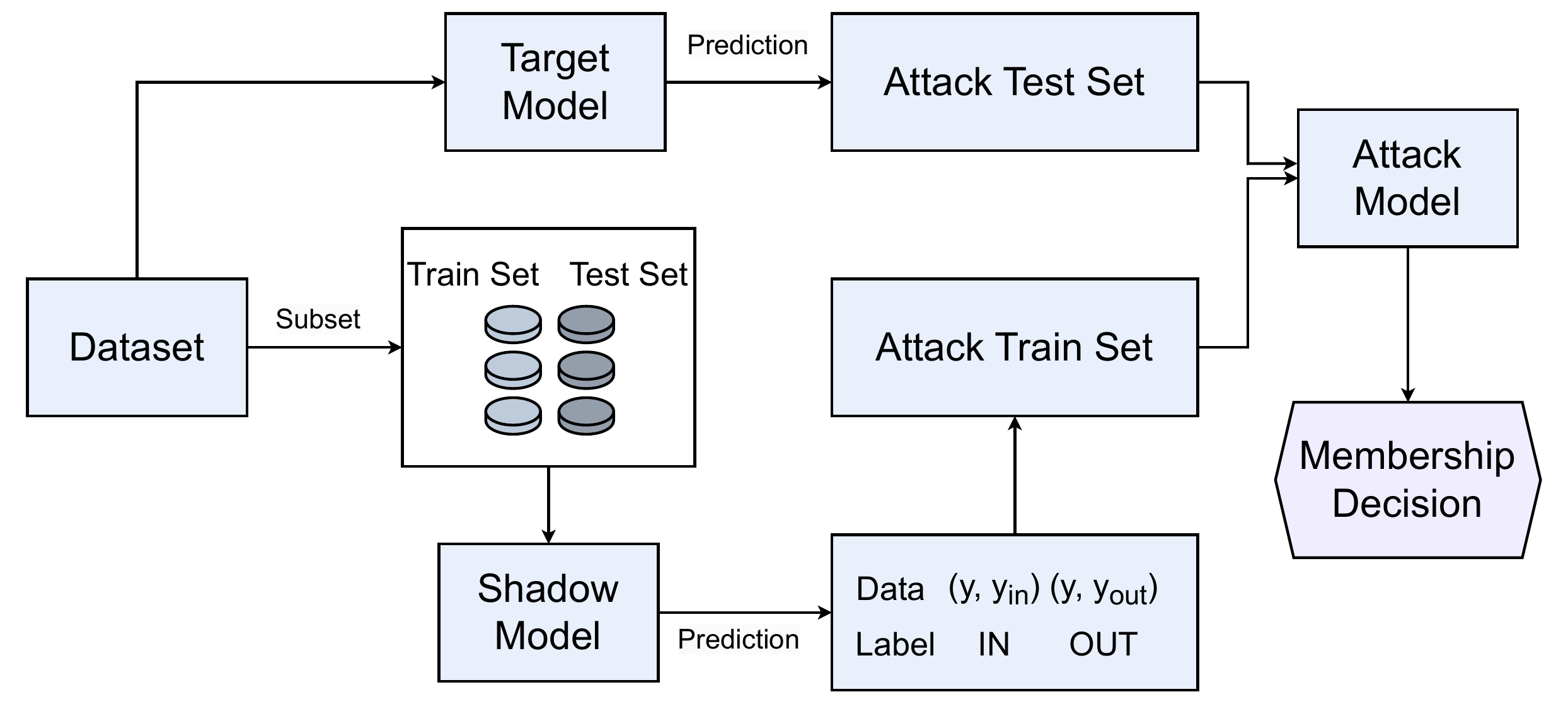}
    \caption{Membership Inference Attack (MIA) Framework}
    \label{fig:mia_24}
\end{figure}
\noindent
MIAs aim to determine whether a given data point was used in a model's training set. These attacks exploit observable discrepancies in model behavior (such as confidence scores or output distributions) between training and non-training samples. Typically, models tend to assign higher confidence or produce more consistent outputs for training samples due to overfitting, which MIAs attempt to detect \cite{shokri}.

The attack is structured as a binary classification task, where each prediction is labeled as either a member (IN) or a non-member (OUT) based on the model’s outputs~\cite{shokri2017membership}. Commonly used features for training the attack model include logits~\cite{shokri}, loss values~\cite{yeom2018privacy}, or internal activations~\cite{nasr2019comprehensive}, depending on the available access to the target. MIAs can be categorized based on the attacker's access to the target model: in black-box settings, only the model outputs (e.g., logits or labels) are available~\cite{salem2019ml}, whereas in white-box settings, internal states such as activations or gradients are accessible~\cite{sablayrolles2019white}. The effectiveness of MIA attacks is often tied to the model’s generalization gap, with larger discrepancies between training and test performance typically indicating weaker resilience to membership inference~\cite{yeom2018privacy}.

In our experiments, we adopt a MIA framework consisting of a \textit{target model} and a \textit{shadow model} as shown in the Figure~\ref{fig:mia_24}, both  both implemented using the same architecture— either SNNConvNet or SNNFCNet (Table~\ref{tab:snnconvnet_arch} and ~\ref{tab:snnfcnet_arch}) and training configuration. The \textit{target model} is trained on the full dataset, while the \textit{shadow model} is trained on a disjoint 80\% subset to replicate the target’s behavior from the attacker’s perspective. The \textit{shadow model} is used to simulate how the \textit{target model} would behave when predicting on member and non-member samples. It is trained on a randomly selected 20\% subset of the dataset, disjoint from the \textit{target model's} training data. To construct the attack dataset, we use the \textit{shadow model's} outputs: predictions on the \textit{shadow model's} own training data are labeled as members (IN), while predictions on its test data are labeled as non-members (OUT). For each input sample passed through the \textit{shadow model}, we record the membrane potentials at the final timestep. These membrane potentials, representing the model’s output activations just before prediction, are used as feature vectors for training the attack model. A support vector machine (SVM)~\cite{joachims1998making} with a radial basis function (RBF) kernel is used as the attack model. Once trained, this attack model is applied to the \textit{target model's} predictions to infer whether an input belongs to the target's training set. We evaluate attack performance using the ROC-AUC score, which quantifies the degree of privacy leakage under different temporal dynamics. An ROC-AUC of 50\% signifies optimal privacy, indicating that the attack model's performance is equivalent to random guessing.

\vspace{-0.5em}
\subsection{Transfer Learning}
\noindent
Transfer learning is a learning paradigm in which knowledge acquired from one task is leveraged to improve performance on a different but related task, often by retaining learned parameters and adapting only a subset of the model~\cite{pan2010survey}.. This approach is valuable when the target dataset is limited in size or presents greater variability or noise, making it more difficult to train models from scratch. A model is initialized with parameters learned from a related source task, enabling the reuse of low-level feature detectors and structural priors. This can accelerate convergence, reduce computational cost, and improve generalization to unseen data~\cite{yosinski2014transfer}. In our study, transfer learning is used to investigate whether spike-based temporal dynamics provide reusable representations across datasets, offering insights into the adaptability and robustness of temporal features in SNNs.

We evaluate this by pretraining the spiking convolutional model~\ref{tab:snnconvnet_arch} on MNIST and fine-tuning it on Fashion-MNIST using the same temporal dynamics. The convolutional layers are frozen after pretraining, while the fully connected layers are retrained on the new dataset. This design isolates the effect of temporal encoding on transferable feature extraction. Separate \textit{shadow models} are trained under the same protocol to construct attack datasets, and MIAs are conducted on the fine-tuned target model to assess the impact of transfer learning on privacy resilience.
\vspace{-0.7em}
\section{Experimental Setup}

\subsection{Dataset and Model Architecture}
\label{subsec:model-architecture}
\noindent
Our experiments utilize both image-based and tabular datasets. For image classification, we used MNIST~\cite{lecun2010mnist}, Fashion-MNIST (FMNIST)~\cite{fashion}, and CIFAR-10~\cite{cifar10}, each consisting of grayscale or RGB images with varying resolutions. To ensure architectural consistency across experiments, all images were resized to $32 \times 32$ pixels. Pixel intensities were normalized to the $[0,1]$ range to support stable spiking activity under temporal encoding. For tabular classification, we used the UCI Breast Cancer dataset~\cite{misc_breast_cancer_14} and the Iris dataset~\cite{iris_53}, which contain 30 and 4 numerical features, respectively. All tabular features were standardized using z-score normalization, ensuring zero mean and unit variance across each feature dimension. The processed datasets were subsequently fed into the spiking models for training and evaluation.

All models are implemented using the \texttt{snnTorch} library~\cite{snntorch}, a PyTorch-based framework for training SNNs, and are executed on an NVIDIA A100 GPU (40GB)\cite{nvidia2020a100}. For image datasets (MNIST, Fashion-MNIST and CIFAR-10), we adopt SNNConvNet architecture (Table \ref{tab:snnconvnet_arch}), while for tabular datasets (Iris and Breast Cancer), we employ SNNFCNet (Table~\ref{tab:snnfcnet_arch}). In these architectures, $K$ denotes the number of output classes (e.g., 10 for CIFAR-10, 3 for Iris), and $F$ denotes the number of input features for tabular datasets (e.g., 30 for Breast Cancer, 4 for Iris).

In both models, we use LIF neurons with a fixed decay constant of $\beta = 0.95$. The simulations run for $T = 100$ discrete timesteps. All models are trained using the Adam optimizer~\cite{kingma2015adam} with a learning rate of 0.001 and batch size of 128. The output layer size is dataset-dependent (e.g., 10 for CIFAR-10, 3 for Iris). The SNNConvNet uses convolutional layers followed by LIF activations and max pooling to extract spatiotemporal features from spike inputs, while SNNFCNet directly processes temporally encoded tabular features.


\begin{table*}[t]
\centering
\begin{minipage}{0.48\linewidth}
\centering
\caption{SNNConvNet Architecture for Image Datasets}
\label{tab:snnconvnet_arch}
\begin{tabular}{lccc}
\toprule
\textbf{Layer} & \textbf{Index} & \textbf{Kernel / Pool} & \textbf{Output Shape} ($C, H, W$) \\
\midrule
Input   & -- & --       & $(C, 32, 32)$ \\
Conv2D  & 0  & $3 \times 3$ & $(32, 32, 32)$ \\
MaxPool & 1  & $2 \times 2$ & $(32, 16, 16)$ \\
Conv2D  & 2  & $3 \times 3$ & $(64, 16, 16)$ \\
MaxPool & 3  & $2 \times 2$ & $(64, 8, 8)$ \\
Flatten & 4  & --       & $4096$ \\
Linear  & 5  & --       & $1000$ \\
LIF     & 6  & --       & $1000$ \\
Linear  & 7  & --       & $K$ \\
LIF     & 8  & --       & $K$ \\
\bottomrule
\end{tabular}
\end{minipage}
\hfill
\begin{minipage}{0.48\linewidth}
\centering
\caption{SNNFCNet Architecture for Tabular Datasets}
\label{tab:snnfcnet_arch}
\begin{tabular}{lcc}
\toprule
\textbf{Layer} & \textbf{Index} &  \textbf{Output Shape} \\
\midrule
Input   & --  & $F$ \\
Linear  & 0   & $1000$ \\
LIF     & 1   & $1000$ \\
Linear  & 2   & $K$ \\
LIF     & 3   & $K$ \\
\bottomrule
\end{tabular}
\end{minipage}
\end{table*}

\vspace{-0.4em}
\subsection{Temporal Dynamics Implementation}
\label{subsec:temporal_implementation}
\noindent

We apply input-level temporal transformations to convert static normalized inputs into dynamic spike representations suitable for spiking network processing. Each transformation generates a spatiotemporal tensor of shape $[T, B, C, H, W]$, where $T$ is the number of discrete time steps (set to 100 in our experiments,) and $B, C, H, W$ correspond to the batch size, channels, height, and width, respectively. In additional to standard rate-based encoding, we investigate two distinct temporal dynamics: Poisson-Burst and Delayed-Burst, described below.

\paragraph{Poisson-Burst Dynamics}
In the Poisson-Burst mechanism, the time axis is divided into non-overlapping intervals of fixed length $\tau$ time steps. Within each interval, a Bernoulli trial is conducted to determine whether a burst occurs, with success probability modulated by the input intensity $x_i$:
\begin{equation}
\Pr(\text{burst in interval}) = f(x_i),
\end{equation}
where $f(x_i)$ is a monotonic function mapping normalized input values to probabilities.

If a burst is triggered, the number of spikes $n_i$ within the burst is sampled from a Poisson distribution:
\begin{equation}
n_i \sim \text{Poisson}(\lambda),
\label{eq:poisson}
\end{equation}
where $\lambda$ controls the mean burst size. Spikes within the burst are placed consecutively with a configurable inter-spike interval (ISI) along the time dimension.

In our experiments, the burst interval $\tau$ was set to \textbf{10} time steps, the Poisson mean $\lambda$ to \textbf{3.0}, and the $\text{ISI}$ was fixed at 1 time step.

\paragraph{Delayed-Burst Dynamics}

In the Delayed-Burst mechanism, each input feature generates exactly one burst during the full time window. The onset delay before the burst $d_i$ for the burst is determined probabilistically based on the input intensity. Specifically, the delay is sampled from a Geometric distribution~\cite{philippou1983generalized}, where the expected value of the delay is inversely related to the input magnitude:
\begin{equation}
\mathbb{E}[d_i] = (1 - x_i) \times (T - m),
\end{equation}
where $x_i$ is the normalized input, $T$ is the total number of time steps, and $m$ is a small margin used to avoid boundary effects. Due to the characteristics of the Geometric distribution, when the input intensity is low, the variance of the sampled delay can become excessively large. To address this, we apply a clipping mechanism that constrains the sampled delay within an interval centered around its expected value, where the width of the interval is controlled by a clipping ratio parameter.

Following the delay, a burst of spikes is generated, where the number of spikes is sampled from a Poisson distribution with mean $\lambda$ (Equation ~\ref{eq:poisson}.) Spikes within the burst are placed at consecutive time steps with a fixed ISI of one time step.
 
For Delayed-Burst dynamics, we set the Poisson mean\textbf{ \textbf{$\lambda$}} to \textbf{4.0} and applied a clipping ratio of 0.2 around the expected delay. 

\noindent Parameter values for all temporal dynamics were selected through empirical tuning and visual inspection of spike patterns to ensure interpretable variability across inputs while maintaining training stability.


\vspace{-0.5em}
\section{Results}

\subsection{Model Performance}
\label{sec:model_performance}
\begin{figure*}[ht]
    \centering
    \includegraphics[width=.82\linewidth, height=0.34\linewidth, trim=10pt 120pt 0pt 0pt]{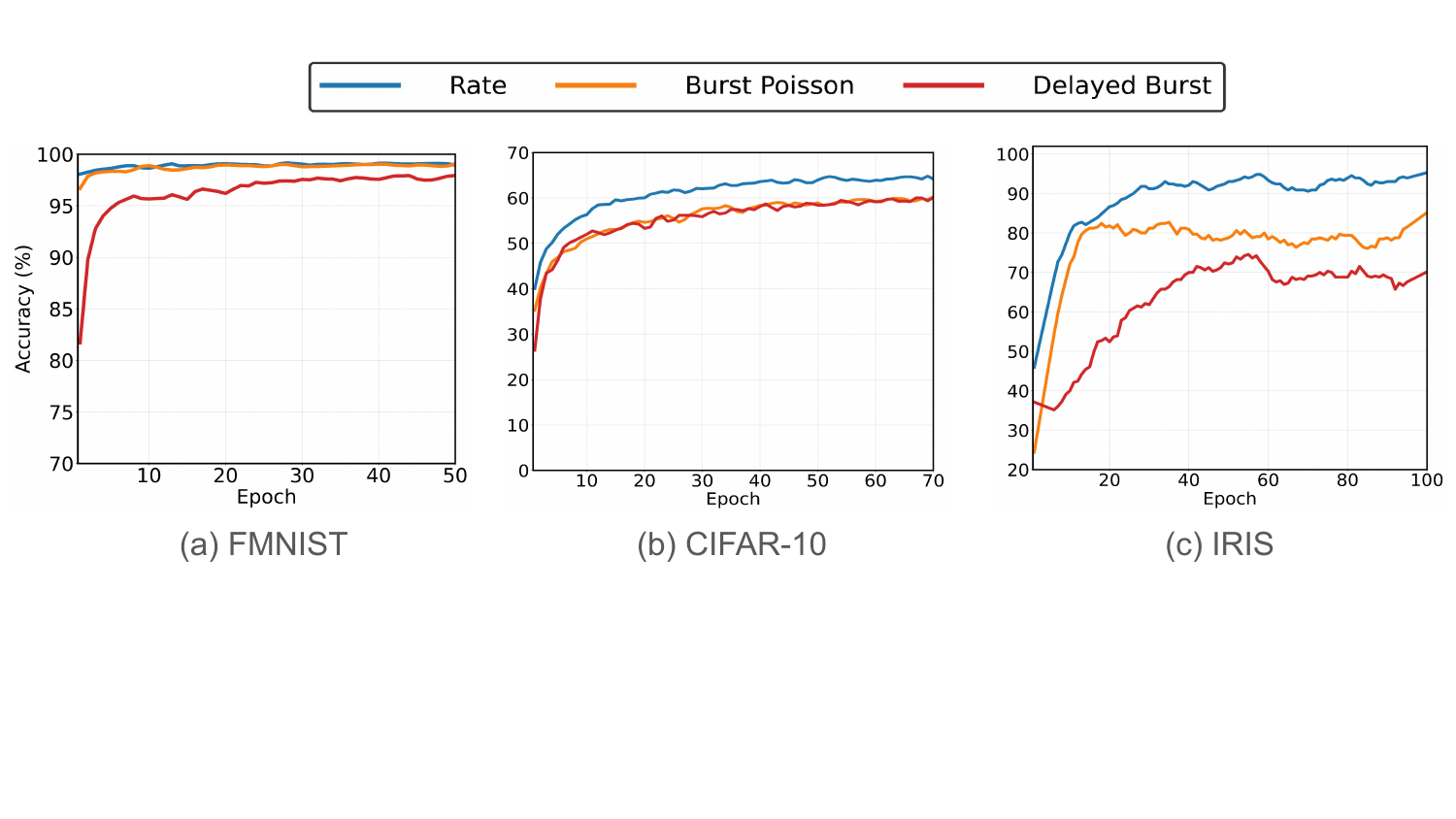}
    \caption{Impact of different temporal spike dynamics on model performance across (a) Fashion-MNIST, (b) CIFAR-10 and (c) Iris datasets.} 
    \label{fig:acc}
\end{figure*}

\begin{figure*}[ht]
    \centering
    \includegraphics[width=.78\linewidth, height=0.31\linewidth, trim=10pt 115pt 0pt 0pt]{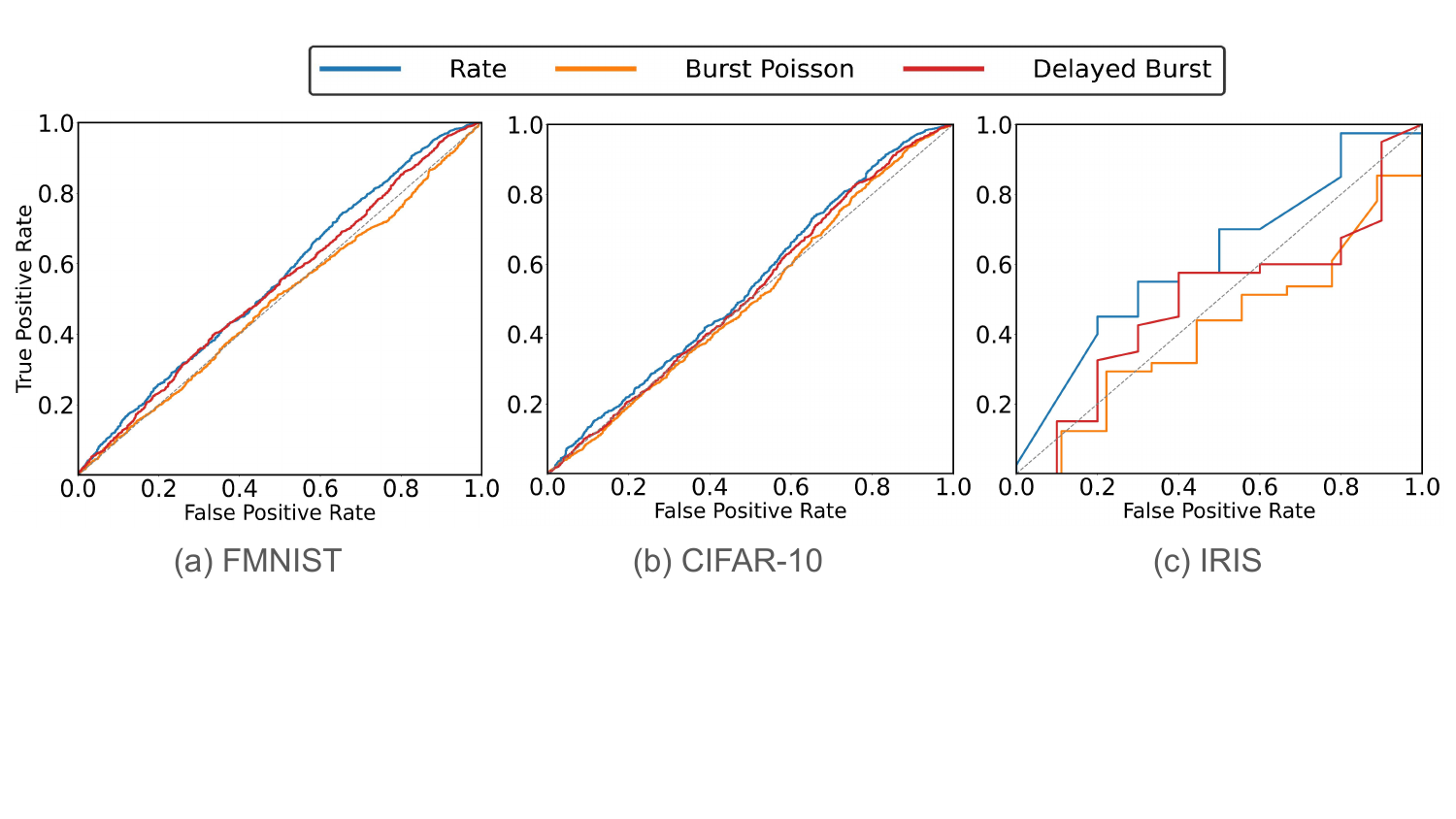}
    \caption{ROC curves illustrating the effect of different temporal spike dynamics on MIA resilience across (a) FMNIST, (b) CIFAR-10, and (c) Iris datasets.}

    \label{fig:priv}
\end{figure*}

\noindent
Standard rate-based spike dynamics serves as the baseline throughout our experiments and consistently delivers the highest test accuracy across datasets as shown in Figure~\ref{fig:acc} and Table~\ref{tab:snn_mia_resilience}. However, the margin by which it outperforms more biologically inspired variants is often narrow, especially on simpler datasets. For example, Poisson-Burst dynamics shows relatively small performance drops on vision datasets like MNIST and FMNIST, with differences under 1.5\%. In contrast, spike dynamics that introduce spike timing delays (Delayed-Burst) exhibit more noticeable performance drops, particularly on CIFAR-10, where Delayed-Burst falls behind the Rate baseline by approximately 7\%. 

The tabular datasets show a similar pattern but with slightly higher variation. Poisson-Burst dynamics retain relatively competitive performance, with drops typically under 3\%, while Delayed-Burst introduces steeper degradation. For example, in Iris, Delayed-Burst leads to a drop of over 10\% in test accuracy compared to standard rate based dynamics. Breast Cancer sees a smaller but consistent degradation of around 4–5\%. These results suggest that spike dynamics preserving tight alignment between spike timing and input magnitude (such as Rate and Poisson-Burst) are better suited for standard SNN architectures. In contrast, by delaying spikes based on input strength, Delayed-Burst dynamics disrupt the expected timing of input features which makes it harder for the network’s feature detectors to recognize input patterns reliably, leading to diminished classification accuracy.

\begin{table}[ht!]
\centering
\caption{Impact of temporal spike dynamics on Accuracy and MIA AUC across MNIST, FMNIST, CIFAR-10, Iris, and Breast Cancer datasets.}
\label{tab:snn_mia_resilience}
\renewcommand{\arraystretch}{1.1}
\setlength{\tabcolsep}{4pt}
\begin{tabular}{p{1.6cm} p{1.9cm} p{1.2cm} p{1cm} p{1.5cm}}
\toprule
\textbf{Dataset} & \textbf{Spike Dynamics} & \textbf{Train Acc} & \textbf{Test Acc} & \textbf{Attack AUC} \\
\midrule
\multirow{3}{*}{\textbf{MNIST}}        
    & Rate              & 99.97\% & 99.26\% & \textbf{0.525(±0.010)} \\
    & Poisson\_Burst    & 99.89\% & 99.09\% & \textbf{0.519(±0.012)} \\
    & Delayed\_Burst    & 98.75\% & 98.04\% & \textbf{0.511(±0.019)} \\    
\midrule
\multirow{3}{*}{\textbf{F-MNIST}}      
    & Rate              & 99.06\% & 90.61\% & \textbf{0.543(±0.048)} \\
    & Poisson\_Burst    & 93.10\% & 88.99\% & \textbf{0.511(±0.012)} \\
    & Delayed\_Burst    & 91.89\% & 89.41\% & \textbf{0.511(±0.013)} \\ 
\midrule
\multirow{3}{*}{\textbf{CIFAR-10}}     
    & Rate              & 73.46\% & 64.85\% & \textbf{0.541(±0.016)} \\
    & Poisson\_Burst    & 65.04\% & 60.42\% & \textbf{0.512(±0.017)} \\
    & Delayed\_Burst    & 62.75\% & 60.27\% & \textbf{0.525(±0.016)} \\ 
\midrule
\multirow{3}{*}{\textbf{Iris}}         
    & Rate              & 97.04\% & 96.67\% & \textbf{0.597(±0.023)} \\
    & Poisson\_Burst    & 96.67\% & 93.33\% & \textbf{0.435(±0.056)} \\
    & Delayed\_Burst    & 93.33\% & 83.33\% & \textbf{0.468(±0.096)} \\ 
\midrule
\multirow{3}{*}{\textbf{Breast Cancer}} 
    & Rate              & 98.90\% & 96.49\% & \textbf{0.545(±0.003)} \\
    & Poisson\_Burst    & 98.24\% & 97.37\% & \textbf{0.531(±0.003)} \\
    & Delayed\_Burst    & 97.67\% & 97.37\% & \textbf{0.489(±0.017)} \\ 
\bottomrule
\end{tabular}
\end{table}
\vspace{-0.37em}
\subsection{Privacy Preservation}
\vspace{-0.37em}
\noindent

Among all datasets, standard rate-based spike dynamics consistently yield the highest attack AUCs, indicating lower resiliency to MIAs. Figure~\ref{fig:priv} and Table~\ref{tab:snn_mia_resilience} report the AUC values across different spike dynamics. Poisson-Burst and Delayed-Burst achieve lower AUCs, suggesting improved privacy. For example, in Breast Cancer, the attack AUC drops from 0.545 (Rate), which is 4.5\% above random guessing, to 0.489 (Delayed-Burst), which is 1.1\% below—yielding a 3.4\% improvement in privacy. Similar privacy gains of 2--5\% are observed in MNIST, FMNIST, and CIFAR-10 relative to the Rate baseline.

These improvements can be attributed to the disruption of output regularities that MIAs typically exploit. Rate-based spike trains produce highly consistent activation profiles tightly correlated with input magnitude, providing strong signals for attackers. In contrast, the temporal variability introduced by Poisson-Burst and Delayed-Burst diffuses memorization cues which makes it harder to distinguish between members and non-members.

As discussed in Section~\ref{sec:model_performance}, these privacy gains are accompanied by slight drops in classification accuracy, particularly for Delayed-Burst. Poisson-Burst offers a favorable privacy–utility balance, maintaining accuracy within 1--3\% of Rate while reducing attack AUC by up to 5\% across several datasets.
\begin{figure*}[ht]
    \centering
    \includegraphics[width=.75
\linewidth, height=0.31\linewidth, trim=30pt 110pt 0pt 0pt]{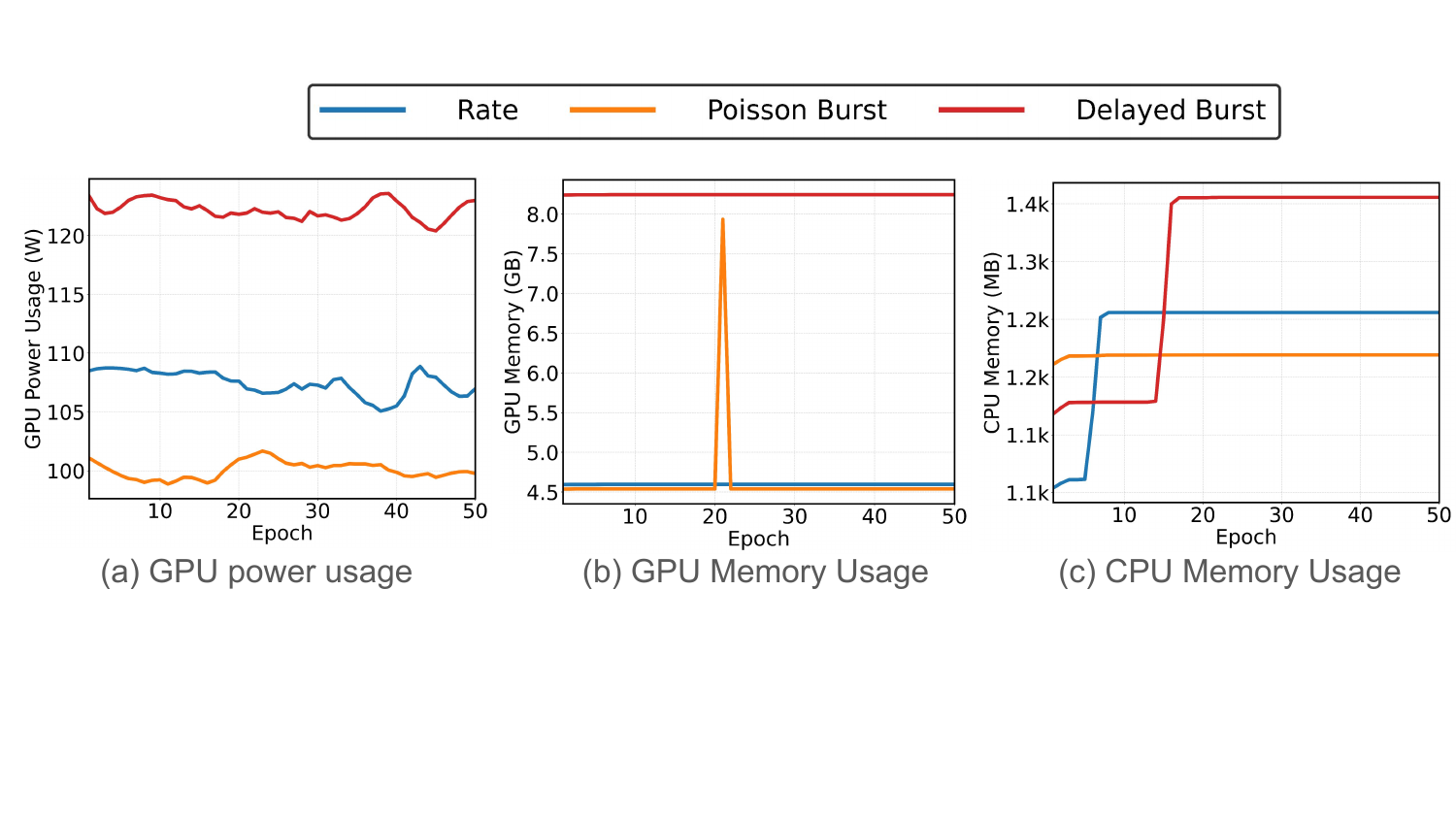}
    \caption{Computational cost on MNIST across temporal spike dynamics: (a) GPU power, (b) GPU memory, (c) CPU memory.} 
    \label{fig:sys}
\end{figure*}
\vspace{-0.4em}
\subsection{Computational Efficiency}

\noindent
To assess system-level cost, we measured GPU power, GPU memory, and CPU memory usage during MNIST training (Figure~\ref{fig:sys}). These metrics capture the practical deployment feasibility of different temporal spike dynamics.

Poisson-Burst exhibits the most efficient behavior, achieving the lowest GPU power and memory usage while maintaining competitive CPU memory consumption. In comparison, Rate dynamics are efficient but slightly more resource-intensive. Delayed-Burst introduces a notable increase in CPU memory usage due to the buffering of delayed spike activations.

These results indicate that system-level overhead is influenced not only by the number of spikes but also by their temporal organization. Regular and moderately stochastic spike patterns, as in Poisson-Burst, distribute computational load more evenly across time, leading to lower peak resource utilization. In contrast, delayed spike dynamics, despite not significantly burdening the GPU (Figure~\ref{fig:sys}a), introduce CPU-side inefficiencies due to their temporal dispersion( Figure~\ref{fig:sys}c). Thus, optimizing for both privacy and efficiency in SNNs requires careful consideration of how spike timing structures impact hardware resource behavior.

\vspace{-0.6em}
\subsection{Transferability}

\begin{table}[ht!]
\centering
\caption{Transferability of Temporal Dynamics from MNIST to FMNIST.}
\label{tab:transfer}
\renewcommand{\arraystretch}{1.1}
\setlength{\tabcolsep}{4pt}
\begin{tabular}{lcc|cc}
\toprule
\multirow{2}{*}{\textbf{Temporal Dynamics}} &
\multicolumn{2}{c|}{\textbf{FMNIST (Scratch)}} & 
\multicolumn{2}{c}{\textbf{FMNIST (Transfer)}} \\
\cmidrule(lr){2-3} \cmidrule(lr){4-5}
& \textbf{Acc. (\%)} & \textbf{AUC} & \textbf{Acc. (\%)} & \textbf{AUC} \\
\midrule
Rate            & 90.61 & \textbf{0.543} & 89.61 & \textbf{0.518} \\
Poisson-Burst   & 88.99 & \textbf{0.511} & 88.16 & \textbf{0.506} \\
Delayed-Burst   & 89.41 & \textbf{0.511} & 86.45 & \textbf{0.503} \\
\bottomrule
\end{tabular}
\end{table}
\vspace{-0.7em}
\noindent
To evaluate whether temporal spike dynamics retain their representational utility across tasks, we conducted a transfer learning experiment by pretraining model~\ref{tab:snnconvnet_arch} on MNIST and fine-tuning the pretrained model on Fashion-MNIST using the same spike dynamics. All dynamics retained high classification accuracy after transfer. Rate-based spike dynamics showed the most consistent performance, maintaining 89.61\% post-transfer compared to 90.61\% when trained from scratch, as shown in Table~\ref{tab:transfer}. Poisson-Burst dynamics similarly exhibited minimal degradation (less than 1\%), indicating that stochastic yet regular burst patterns are robust to task shift. In contrast, Delayed-Burst dynamics showed greater sensitivity to transfer, dropping by about 3\% in test accuracy. This suggests that latency-based representations may be more tightly coupled to the training distribution and less transferable across domains.

From a privacy perspective, all spike dynamics exhibit improved MIA resilience following transfer learning, as evidenced by lower attack AUCs compared to their corresponding models trained from scratch. Although Rate dynamics show the largest relative drop in attack AUC (4.6\%), they still remain the most vulnerable post-transfer, with a final AUC higher than Poisson-Burst and Delayed-Burst. Poisson-Burst and Delayed-Burst achieve smaller relative reductions (approximately 1.0\% and 1.6\%, respectively) but end with lower absolute AUCs, indicating stronger overall privacy protection after domain shift. These results suggest that transfer learning reduces membership inference vulnerability across all spike dynamics, with spike timing variability being associated with further improvements in privacy resilience.

\vspace{-0.4em}
\section{Conclusion}
\noindent
While SNNs offer temporal coding advantages and energy-efficient computation, their widespread reliance on simplified neuron models such as LIF constrains their ability to replicate the rich temporal dynamics observed in biological systems. Beyond biological fidelity, the lack of rich temporal diversity in typical SNNs has been hypothesized to limit robustness, generalization, privacy resilience, and scalability in practical applications. Motivated by these limitations, we introduced Poisson-Burst and Delayed-Burst spike dynamics, which add stochastic variability in burst size and modulate spike timing based on input strength, while preserving standard LIF neuron dynamics. Our findings demonstrate that moderate spike timing variability, as introduced through Poisson-Burst dynamics, reduces memorization and computational overhead by distributing spike activity more evenly across time, thereby improving privacy resilience and system efficiency with minimal loss in accuracy. In contrast, excessive irregularity from input-dependent delays in Delayed-Burst dynamics disrupts the temporal coherence needed for effective feature extraction, resulting in greater accuracy degradation despite stronger privacy gains. These results indicate that the specific nature and degree of spike timing variability, beyond overall spike rate alone, can influence the trade-offs between predictive performance, privacy protection, generalization and resource demands in SNNs.

Looking ahead, an important extension is to move beyond fixed spike dynamics toward adaptive temporal strategies, where patterns such as burst size and onset latency are optimized during training based on task demands. Further scaling these insights to architectures with richer neuron models, and supporting them with energy-efficient beyond-CMOS devices that can reproduce diverse firing behaviors, could help advance the development of scalable, privacy-resilient SNNs. These advances point toward scalable, hardware-efficient SNNs that combine robustness, privacy protection, and adaptivity for complex, event-driven computational tasks.

\vspace{-0.5em}

\end{document}